\documentclass{article}

\usepackage[square,numbers]{natbib}

\usepackage[final]{nips_2017}

\usepackage[utf8]{inputenc} 
\usepackage[T1]{fontenc}    
\usepackage{hyperref}       
\usepackage{url}            
\usepackage{booktabs}       
\usepackage{amsfonts}       
\usepackage{nicefrac}       
\usepackage{microtype}      
\usepackage{amsmath}
\usepackage[pdftex]{graphicx} 
\graphicspath{{figures/}}
\usepackage{float}
\floatstyle{plaintop}
\restylefloat{table}

\expandafter\def\expandafter\normalsize\expandafter{%
    \normalsize
    \setlength\abovedisplayskip{6pt}
    \setlength\belowdisplayskip{3pt}
    \setlength\abovedisplayshortskip{6pt}
    \setlength\belowdisplayshortskip{3pt}
}

\usepackage{tabu}

\usepackage{titlesec}
\titlespacing*{\section}{0pt}{0.7\baselineskip}{0.7\baselineskip}
\titlespacing*{\subsection}{0pt}{0.7\baselineskip}{0.7\baselineskip}

\usepackage[tableposition = top]{caption}

\title{Learning Loss Functions for Semi-supervised Learning via Discriminative Adversarial Networks}

\author{
  Cicero Nogueira dos Santos, Kahini Wadhawan, Bowen Zhou \\
  AI Foundations, IBM Research\\
  IBM T.J Watson Research Center\\
  \texttt{\{cicerons,zhou\}@us.ibm.com}, \texttt{kahini.wadhawan@ibm.com} \\
}

\begin{document}

\maketitle

\begin{abstract}
We propose discriminative adversarial networks (DAN) for semi-supervised learning and loss function learning.
Our DAN approach builds upon generative adversarial networks (GANs) and conditional GANs but includes the key differentiator of using two discriminators instead of a generator and a discriminator.
DAN can be seen as a framework to learn loss functions for predictors that also implements semi-supervised learning in a straightforward manner. 
We propose instantiations of DAN for two different prediction tasks: classification and ranking.
Our experimental results on three datasets of different tasks
demonstrate that DAN is a promising framework for both semi-supervised learning and learning loss functions for predictors.
For all tasks, the semi-supervised capability of DAN can significantly boost the predictor performance for small labeled sets with minor architecture changes across tasks.
Moreover, the loss functions automatically learned by DANs are very competitive and usually outperform the standard pairwise and negative log-likelihood loss functions for both semi-supervised and supervised learning.
\end{abstract}

\section{Introduction}
One of the challenges in developing semi-supervised learning (SSL) algorithms is to define a loss (cost) function that handles both labeled and unlabeled data.
Many SSL methods work by changing the original loss function to include an additional term that deals with the unlabeled data \cite{zhu2005semi,miyato2015distributional,SajjadiJT16}.
Recent advances in generative models have allowed the development of successful approaches that perform SSL while doing data generation,
which allows the use of unlabeled data in more flexible ways.  
The two main families of successful generative approaches are based on variational autoencoders (VAE) \cite{kingma2013auto} and generative adversarial networks  \cite{goodfellow2014generative}.
Most GAN-based SSL approaches change the loss function of the discriminator to combine a supervised loss (e.g. negative log likelihood with respect to the ground truth labels) with the unsupervised loss normally used in the discriminator \cite{springenbergICLR16}.
While VAE-based SSL approaches have achieved good results for tasks in both computer vision \cite{kingma2014semi,maaloe16} and natural language processing domains \cite{xu2017variational,yangHSB17},
GAN-based SSL have primarily targeted tasks from the computer vision domain  \cite{salimans2016improved,odena2016semi,dumoulinICLR17,kumar2017}. The main reason is that applying GANs to discrete data generation problems, e.g. natural language generation, is difficult because the generator network in GAN is designed to be able to adjust the output continuously, which does not (naturally) work on discrete data generation.

In this paper, we propose discriminative adversarial networks (DAN) for SSL and loss function learning.
DAN builds upon GAN and conditional GAN but includes the key differentiator of using two discriminators instead of a generator and a discriminator.
The first discriminator (the \emph{predictor} $P$) produces the prediction $y$ given a data point $x$, and the second discriminator (the \emph{judge} $J$) takes in a pair $(x, y)$ and judges if it is a {\it predicted label} pair or {\it human labeled} pair.
While GAN can be seen as a method that implicitly learns loss functions for generative models,
DAN can be seen as a method that learns loss functions for predictors.
The main benefits of DAN are:
\begin{itemize}
\item The predictor $P$ does not use information from labels, therefore 
unlabeled data can be used in a transparent way;
\item We do not need to manually define a loss function that handles both labeled and unlabeled data, the judge $J$ implicitly learns the loss function used to optimize $P$;
\item Different from VAE and GAN-base SSL approaches, in DAN we do not have to perform data generation. This allows the application of SSL using adversarial networks for natural language processing (NLP) sidestepping troubled discrete data generation;
\item Prediction problems with complex/structured outputs can benefit from DAN's implicit loss function learning capability. This is important because for many structured prediction problems such as ranking and coreference resolution, researchers normally use surrogate loss functions since the best loss function for the problem is too expensive to compute or, in some cases, because a good loss function is not even known.
\end{itemize}


We have applied DAN for two different NLP tasks, namely, answer sentence selection (ranking) and text classification.
We have proposed simple but effective DAN architectures for both tasks.
We have also introduced new scoring functions for the judge network that makes the training more stable.
Our experimental results demonstrate that:
(1) DAN can boost the performance when using a small number of labeled samples;
(2) the loss functions automatically learned by DAN outperform standard-pairwise and negative log-likelihood loss functions for the semi-supervised setup, and is also very competitive (and many times better) in the supervised setting.


The remaining of this paper is organized as follows. In Sec. \ref{sec:methods} we give a brief overview of GANs and conditional GANs, followed by a detailed description of our  proposed approach. In Sec. \ref{sec:relatedwork} we discuss the related work. Sec. \ref{sec:experiments} details our experimental setup and results. Finally, Sec. \ref{sec:conclusion} brings some concluding remarks.


\section{Methods}
\label{sec:methods}

In this section, we present the DAN framework and detail its instantiation for two different tasks: classification and ranking.  For the benefit of the method presentation, we first describe GANs and conditional GANs approaches.

\subsection{Generative Adversarial Nets}
Generative adversarial networks are an effective approach for training generative models \cite{goodfellow2014generative}. 
The GAN framework normally comprises two ``adversarial'' networks: a generative net G that ``learns'' the data distribution, 
and a discriminative net D that estimates the probability that a sample came from the real data distribution rather than generated by G. 
In order to learn a generator distribution $p_{g}$ over data $x$, the generator builds a mapping function from a prior noise distribution $p_{z}(z)$ to the data space as $G(z;\theta_{g})$. 
The discriminator receives as input a data point $x$ and outputs a single scalar, $D(x;\theta_{d})$, which represents the probability that $x$ came from training data rather than $p_{g}$ .
 
G and D are trained simultaneously by adjusting parameters for G to minimize $log(1 - D(G(z))$ and adjusting parameters for D to minimize $logD(x)$, as if they are following a two-player min-max game with the following value function $V(G, D)$:
\begin{equation} \label{eq:2.1}
\underset{G}{min} \hspace{1mm} \underset{D}{max}\hspace{1mm} V(D,G) = \mathbb{E}_{x \sim p_{data}(x)} \hspace{1mm} [log D(x)] + \mathbb{E}_{z \sim p_{z}(z)} \hspace{1mm} [log(1-D(G(z)))] 
\end{equation}

\subsection{Conditional Adversarial Nets}
Generative adversarial nets can perform conditional generation if both the generator and discriminator are conditioned on some extra information $y$ \cite{mirza2014conditional}. Normally $y$ is a class label or other type of auxiliary information. The conditioning is performed by feeding $y$ into both the discriminator and generator as an additional input.

In the generator, the prior input noise $p_{z}(z)$ and $y$ are combined in a joint hidden representation. Usually this consists of simply concatenating a vector representation of $y$ to the input vector $z$.
The discriminator receives both $x$ and $y$ as inputs and has to discriminate between real $x,y$ and generated $G(z,y)$.
The objective function of the two-player minimax game can be formulated as follows:
\begin{equation} \label{eq:2.2}
\underset{G}{min} \hspace{1mm} \underset{D}{max}\hspace{1mm} V(D,G) = \mathbb{E}_{x,y \sim p_{data}(x,y)} \hspace{1mm} [log D(x, y)] + \mathbb{E}_{z \sim p_{z}(z),y \sim p_{y}(y)} \hspace{1mm} [log(1-D(G(z, y), y))] 
\end{equation}


\subsection{Discriminative Adversarial Networks}
\begin{figure}[H]
\begin{center}
\includegraphics[width=0.8\linewidth, height=0.3\linewidth]{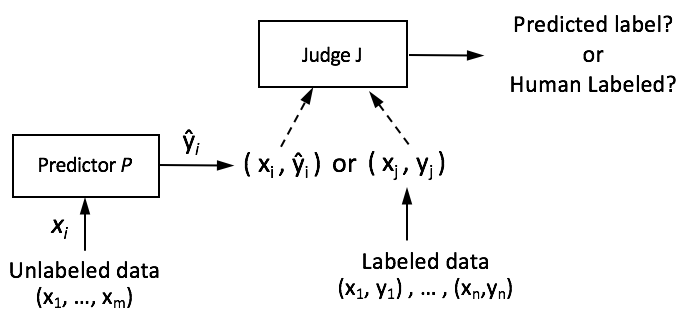}
\caption{DAN framework}
\label{fig:dan-framework}
\end{center}
\end{figure}

We call DAN the adversarial network framework that uses discriminators only.
Here we propose a DAN formulation that allows semi-supervised learning.
In our DAN formulation we use two discriminators: the \emph{Predictor} $P$ and the \emph{Judge} $J$. 
$P$ receives as input a data point $x$ and outputs a prediction $P(x)$. The prediction can be a simple probability distribution over class labels or any other sort of structured predictions such as trees or document rankings.
The Judge network $J$ receives as input a data point $x$ and a label $y$ \footnote{We are using the term \emph{label} in a loose way to also mean any type of structured prediction.} and produces a single scalar, $J(x, y)$, which represents the probability that $x,y$ came from the labeled training data rather than predicted by $P$. Fig. \ref{fig:dan-framework} illustrates the DAN framework. While in conditional GANs the idea is to generate $x$ conditioned on $y$, in DAN we want to predict $y$ conditioned on $x$. 
The min-max game value function V($J$,$P$) becomes: 
\begin{equation} \label{eq:3.1}
\underset{P}{min} \hspace{1mm} \underset{J}{max}\hspace{1mm} V(J,P) = \mathbb{E}_{x,y \sim p_{data}(x,y)} \hspace{1mm} [log J(x, y)] + \mathbb{E}_{x \sim p_{data}(x)} \hspace{1mm} [log(1-J(x, P(x)))]
\end{equation}


An important characteristic in our DAN formulation is that $P$ does not make use of labels, which makes semi-supervised learning straightforward in this framework. This framework also alleviates the need of specifying a loss function for the predictor, the loss function can be learned implicitly by the Judge. The following two subsections describe the instantiation of DAN for two different tasks: classification and ranking.

\subsubsection{DAN for Text Classification}
\begin{figure}[H]
\begin{center}
\includegraphics[width=0.9\linewidth, height=0.4\linewidth]{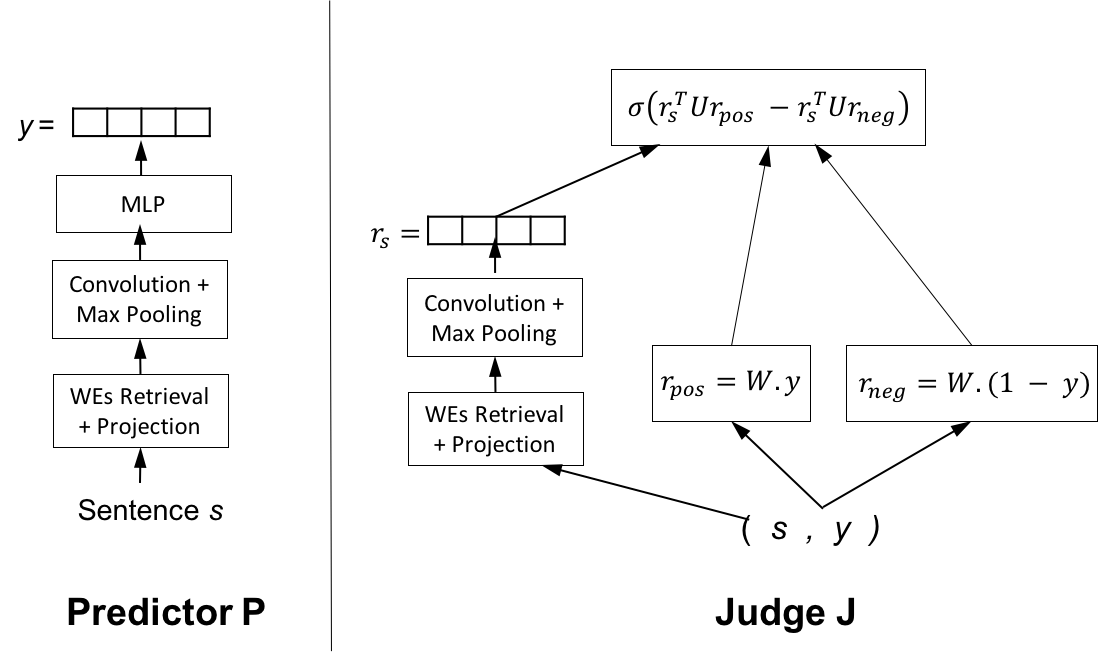}
\caption{DAN Architecture for Text Classification}
\label{fig:pj-text}
\end{center}
\end{figure}

As illustrated in left hand side of Fig. \ref{fig:pj-text}, the Predictor $P$ is a standard CNN-based text classifier that classifies a given sentence $s$ into one of $N$ classes. It takes in sentence $s$ as input and outputs $y$ a probability distribution over $N$ classes. We first retrieve the word embeddings (WEs) and project them using a fully connected layer. Next, a convolutional layer followed by a MLP is used to perform the prediction.

The Judge $J$ takes in a pair ($x$,$y$) consisting of a sentence and its class label, and classifies the pair as being predicted label (fake) or human labeled pair (real).
For the human labeled pairs, $y$ is encoded as the one hot representation of the class label.
The predicted $y$'s is the probability distribution over class labels.
As in the Predictor, we create a representation $r_s$ of the sentence using a convolution.
For Judge as can be noticed in right hand side of Fig. \ref{fig:pj-text},
we create two representations of the class label $y$, $r_{pos}$ and $r_{neg}$ using a embedding matrix $W$.
The representation $r_{pos}$, 
can be seen as the embedding of the positive/correct label.
While the representation $r_{neg}$ can be understood as the average embedding of the negative classes.
The final scoring is done by first measuring the similarity between $r_s$ and $r_{pos}$, and between $r_s$ and $r_{neg}$ using bilinear forms: $r_{s}^{T}Ur_{pos}$ and $r_{s}^{T}Ur_{pos}$, where $U$ is a matrix of learnable parameters.
This type of bilinear similarity measure has been previously used for embedding comparison in \cite{BordesWU14}. 
Next, the difference between the two similarities are passed through the sigmoid function ($\sigma$).
The rationale behind this scoring function is that, if the given label is correct, the representation of the sentence, $r_s$, should be more similar to $r_{pos}$ than to $r_{neg}$.
In our experiments, this scoring approach has shown to be empirically easier to train under the min-max game than concatenating $r_{pos}$ and $r_s$ and giving them as input to a logistic regression (or MLP).
We developed this scoring approach for the ranking task first (next section)   and later realized that it also works well for classification.
\vspace{-6ex}

\subsection{DAN for Answer Selection / Ranking}
In the answer selection task, given a question $q$ and a candidate answer pool $P = (a_{1},a_{2},\cdots ,a_{M})$ for $q$, the goal is to search for and select the candidate answer(s) $a \in P$ that correctly answers $q$. 
This task can be viewed as a ranking problem where the goal is to rank the candidate answers from the best to the worst.
People normally use the following pairwise ranking loss function (hinge loss) when optimizing neural network based rankers: 
$L = \max \{ 0, l - s_{\theta}(q, a^+) + s_{\theta}(q, a^-) \}$,
where $a^+$ is a positive answer, $a^-$ is a negative answer and $l$ is a margin.
However, pairwise ranking loss is known to be suboptimal \cite{Cao:2007:Listwise}.
Our goal on choosing this ranking task is two fold: (1) we believe that the semi-supervised nature of DANs can help to reduce the need of labeled data for answer selection; (2) we believe that DANs can learn a good listwise loss function by taking into consideration the scoring of the whole set of candidate answers.

\begin{figure}[H]
\begin{center}
\includegraphics[width=1\linewidth,height=0.45\linewidth]{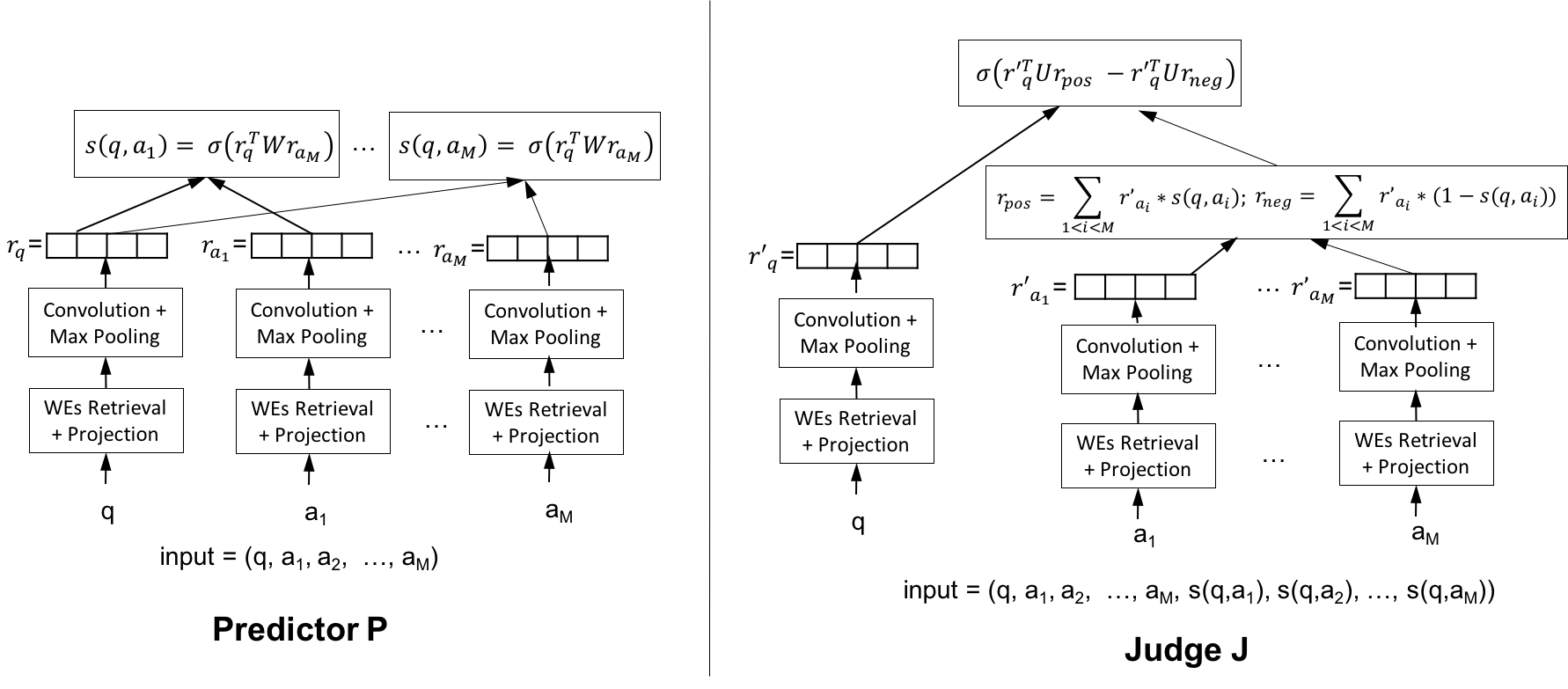}
\caption{DAN Architecture for Answer Selection}
\label{fig:pj-qa}
\end{center}
\end{figure}

As depicted in the left hand side of Fig. \ref{fig:pj-qa}, the Predictor $P$ takes as input the list $(q, a_{1}, a_{2},  \text{…}, a_{M})$ containing a question $q$ and $M$ candidate answers. Pairs of $(q,a_{i})$ are processed in parallel by first generating fixed-length continuous vector representations $r_{q}$ and $r_{a_{i}}$ and then performing the operation $\sigma(r_{q}^{T}Wr_{a_{i}})$,
where $W$ is a matrix of learnable parameters and $\sigma$ is the sigmoid function. Since we are using a sigmoid, note that the score produced by $P$ is a number between 0 and 1.
The parameters of WE projection layer, convolution layer and $W$ are shared among question and all candidate answers.
\vspace{-6ex}
The right hand side of Fig. \ref{fig:pj-qa} details the Judge $J$, which uses a similar architecture as the predictor, except for the scoring function.  There is no parameter sharing between $P$ and $J$. Note that $J$ also receives as input the score for each candidate answer, which means that $J$ performs a listwise scoring.
For the labeled instances, the score for a correct answer is 1 and for an incorrect answer is 0. 

After creating the representation $r'_q$, $r'_{a_{1}}$, ..., $r'_{a_{m}}$, the Judge J uses the scores $s_{a_{1}}$, ..., $s_{a_{m}}$ to compute representations $r_{pos}$ and $r_{neg}$ as follows:
\begin{gather} \label{eq:3.3}
r_{pos} = \sum_{1 < i <= M}{r'_{a_{i}}*s(q,a_{i})} \\
r_{neg} = \sum_{1 < i <= M}{r'_{a_{i}}*(1-s(q,a_{i}))}
\end{gather}
We can think of $r_{pos}$ and $r_{neg}$ as a way to summarize, according to the scores, the similarities and dissimilarities, respectively, between the question and the list of candidate answers.
The final scoring is given by $\sigma(r'^{T}_{q}Ur_{pos} - r'^{T}_{q}Ur_{neg})$. The rationale behind this scoring function is that, if the given list of scores is good, the representation of the question, $r'_q$, should be more similar to $r_{pos}$ than to $r_{neg}$.
As far as we know, this scoring function is novel, and we further extended it for the classification task as presented in the previous section.

\section{Related work} 
\label{sec:relatedwork}
The approach proposed in this work is mainly related to recent works on semi-supervised GANs and conditional GANs.
\citet{springenbergICLR16} proposed a categorical generative adversarial network (CatGAN) which can be used for unsupervised and semi-supervised learning, where the discriminator outputs a distribution over classes and is trained to minimize the predicted entropy for real data and maximize the predicted entropy for fake data. \citet{salimans2016improved} proposed a semi-supervised GAN model in which the discriminator outputs a softmax over classes rather than a probability of real vs. fake. An additional ``generated'' class is used as the target for generated samples. \citet{kumar2017} use the same GAN-like SSL setup proposed in \cite{salimans2016improved},
but use tangents from the generator's mapping to further improve on SSL.
Different from these past works, in DAN we do not perform a generation step, therefore it is easier to apply for discrete data.

Regarding loss function learning using GAN-like approaches, 
\citet{isola2017image} proposed conditional GANs for image-to-image translation problems, and showed that their models not only learns good mappings but also learns a loss function to train the mapping.
\citet{FinnCAL16} presented a connection between GAN-based loss function learning for generative models and cost function learning in reinforcement learning (aka inverse reinforcement learning).
They demonstrated that certain IRL methods are mathematically equivalent to GANs.
While previous work focus on learning loss functions for generative models, in DAN we focus on learning loss functions for discriminative models.

Another recent line of work consists in using adversarial examples \cite{szegedy2013intriguing,goodfellow2014explaining} based on unlabeled data to regularize the training \cite{miyato2015distributional}.
For the NLP domain, the work by \citet{miyato2016virtual} extended the adversarial and virtual adversarial training approaches by adding small perturbations to word embeddings. They report good performance for semi-supervised text classification tasks.
In DAN, instead of adding an extra regularization term to the supervised loss, we implicitly learn the loss function.

Regarding the answer selection task, 
various neural models based on attention mechanisms have been recently proposed \cite{wang2016machine,dos2016attentive,xiong2016dynamic,yang2015wikiqa,dhingra2016gated,sordoni2016iterative,cui2016attention}.
However,
these neural net models only perform well when a large amount of labeled data is available for training.
In contrast, 
our DAN-based answer selection approach is an alternative that can be used when few labeled instances are available.


\section{Experiments}
\label{sec:experiments}
\subsection{Setup and Datasets}
We use two different datasets to perform our answer selection experiments: SelQA \cite{jurczyk2016selqa} and WikiQA \citet{yang2015wikiqa}. Both contain open domain  questions whose answers were extracted from Wikipedia articles.
For both datasets, 
we use the subtask that assumes that there is at least one correct answer for a question.
For the WikiQA, 
the corresponding dataset consists of 873 questions in the training set 
(20,360 question/candidate pairs), 
126 in the dev set (1,130 pairs) and 243 questions in the test set (2,352 pairs).
For SelQA, 
the corresponding dataset contains of 5529 questions in the training set, 
785 in the dev set and 1590 questions in the test set.
SelQA is more than 6 times larger than WikiQA in number of questions.

For the text classification task, 
we use the Stanford Sentiment Tree-bank (SSTb) dataset.
It is a movie review dataset proposed by \citet{socher2013recursive}, which includes fine grained sentiment labels for 215,154 phrases in the parse trees of 11,855 sentences. In our experiments we focus on sentiment prediction of complete sentences only and perform binary classification only.
This dataset,
which is known as SSTB2,
contains 6,920 training sentences,
872 dev. sentences and 1,821 test sentences.

In all experiments, 
we use word embeddings of size 400,
which were pre-trained using the word2vec tool \cite{word2vec2013}.
For the answer selection task we use a dump of Enlgish Wikipedia. 
For sentiment classification,
we pretrain the word embeddings using the IMDB data proposed by \citet{maasEtAl:2011}.

We use the ADAM optimizer \cite{KingmaB14:adam},
and kept the values of most of the hyperparameters fixed for all the experiments. For both the Predictor and the Judge, the word embeddings projection layer has 200 units, the convolutional layer has 400 filters, with context window of sizes 3 and 5 words in the case of answer selection and text classification, respectively. The  
$U$ matrix has dimensionally $\mathbb{R}^{400 \times 400}$.
When training using the full dataset, 
we alternately update $J$ and $P$ one time each. 
We use a learning rate of $\lambda = 0.0005$ for the answer selection task, and $\lambda = 0.0001$ for the text classification task. 
Validation sets are used to perform early stopping.
Normally it is needed less than 50 epochs to achieve the best performance in the validation set.

For the semi-supervised experiments, 
since the set of unlabeled instances is much large than the one of labeled, 
we noticed that we need to update $P$ more frequently than $J$ in order to avoid overfitting $J$.
For better results in the semi-supervised setup, 
we normally update $P$ 10 times after each update of $J$. However, in this case we also had to use a smaller learning rates for $P$ ( $\lambda = 0.00005$) and $J$ ($\lambda = 0.0001$).

For both tasks, 
answer selection and sentiment classification,
we perform semi-supervised experiments where we randomly sample a limited number of labeled instances and use the rest of the dataset as unlabeled data.
In all experiments reported in the next two sections, we repeat the random sampling 10 times and average the results.
Additionally,
in the experiments using the full labeled dataset we repeat the experiments 10 times with different seeds for the random number generator and average the results.  

For both tasks,
we use the term CNN-DAN to refer to the DAN architecture for that respective task (Figs. \ref{fig:pj-text} and \ref{fig:pj-qa}).
However, in the CNN-DAN setup,
the instances presented to $P$ are the exact same instances that appear in the labeled set. 
Therefore, CNN-DAN is basically trying to learn a better loss function using the available labeled data, no semi-supervised learning is performed.
We use the term CNN-DAN$_{unlab.}$ to refer to the DAN setup where we feed $P$ with additional unlabeled data.



\subsection{Answer Selection Results}
\begin{figure}[H]
    \centering
    \begin{minipage}{0.5\textwidth}
        \centering
        \includegraphics[width=1.0\textwidth]{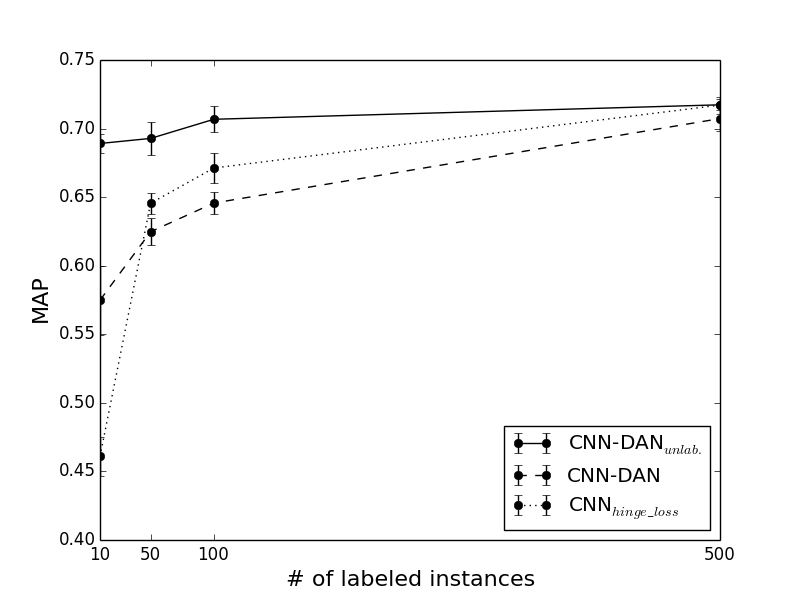} 
        \caption{SelQA}
        \label{fig:selqa}
    \end{minipage}\hfill
    \begin{minipage}{0.5\textwidth}
        \centering
        \includegraphics[width=1.0\textwidth]{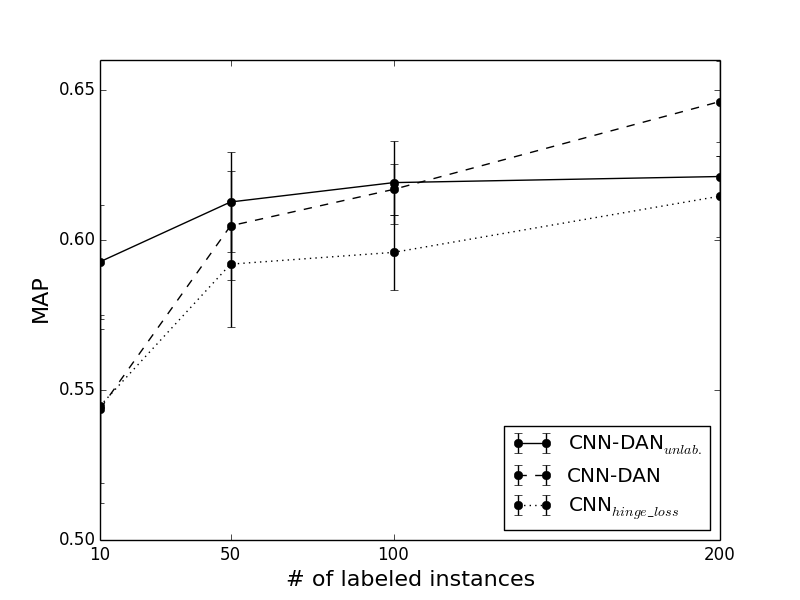} 
        \caption{WikiQA}
        \label{fig:wikiqa}
    \end{minipage}
\end{figure}

In Tables \ref{selqa-results} and \ref{wikiqa-results} we present the experimental results for SelQA and WikiQA, respectively.
We use consolidated ranking metrics to assess the output of the models: mean average precision (MAP), mean reciprocal rank (MRR) and normalized discounted cumulative gain (NDCG).
We present results for CNN-DAN, CNN-DAN$_{unlab.}$ (that uses unlabeled data in $P$) and CNN$_{hinge\_loss}$, 
which is the same CNN-based architecture of the predictor $P$ in our DAN for answer selection (Fig. \ref{fig:pj-qa}), 
but that is trained using the hinge loss function instead of the DAN framework.
We present detailed results for datasets containing a different number of labeled instances: 10, 50 and full dataset.
In Figs. \ref{fig:selqa} and \ref{fig:wikiqa}, 
we also present the MAP for datasets with 100 and 500 labeled instances.

We can see in Figs. \ref{fig:selqa} and \ref{fig:wikiqa} that the semi-supervised DAN, CNN-DAN$_{unlab.}$, gives a significant boost in performance when a small amount of labeled instances is available.
When using 10 labeled instances only (i.e. 10 questions and their respective candidate answers) CNN-DAN$_{unlab.}$ achieves MAP of 0.6891 for the SelQA test set, while the CNN$_{hinge\_loss}$ achieves MAP of 0.4610 only, a difference of approximately 50\%.
A similar behavior can also be seen for the WikiQA dataset, 
where CNN-DAN$_{unlab.}$ consistently has significantly better performance for small labeled sets.
These results are evidence that DAN is a promising approach for semi-supervised learning.

Comparing CNN-DAN, that does not used unlabeled data, 
with CNN$_{hinge\_loss}$ is a reasonable way to check whether the learned loss function is doing better or not than the pairwise hinge loss.
For both datasets, when only 10 data points are available CNN-DAN produces better results than CNN$_{hinge\_loss}$.
For the WikiQA dataset, we can see that the loss function learned by CNN-DAN is doing a better job than the pairwise ranking loss for small and large labeled sets.
When the full dataset is used, 
CNN-DAN achieves an average MAP of 0.6663 while CNN$_{hinge\_loss}$ achieves MAP of 0.6511.

When compared to state-of-the-art results, for SelQA our baseline CNN$_{hinge\_loss}$ outranked the previously best reported result from \cite{jurczyk2016selqa}, which used Attentive Pooling Networks.
For the WikiQA dataset, our CNN-DAN achieves a result comparable to other recently proposed models that use similar CNN architectures \cite{yang2015wikiqa,yin2015abcnn,dos2016attentive}.
\citet{wang2017bilateral} use a model that is way more sophisticated than our one-layer CNN predictor. In our experiments we have used simple architectures in both predictor and judge in order to make it easier to check the real contribution of the proposed approach.
\begin{table}[H]
  \caption{Experimental Results for the SelQA dataset.}
  \label{selqa-results}
  \centering
  \begin{tabu} to \textwidth {lXXXXXXXXX}
    \toprule
    
    \multicolumn{1}{c}{Model} & \multicolumn{3}{c}{10 labeled instances} & \multicolumn{3}{c}{50 labeled instances} & \multicolumn{3}{c}{Full dataset}\\
    \cmidrule{2-10}
                        & MAP & MRR & NDCG & MAP & MRR & NDCG & MAP & MRR & NDCG \\
    \midrule
    CNN$_{hinge\_loss}$   & 0.4610 & 0.4661 & 0.5889 & 0.6455 & 0.6545 & 0.7331 & \textbf{0.8758} & \textbf{0.8812} & \textbf{0.9079}  \\ 
    CNN-DAN              & 0.5749 & 0.5811 & 0.6780 &0.6248 & 0.6332 & 0.7170 & 0.8655 & 0.8730 & 0.9012  \\
    CNN-DAN$_{unlab.}$ & \textbf{0.6891} & \textbf{0.6978} & \textbf{0.7667} & \textbf{0.6928} & \textbf{0.7017} & \textbf{0.7695} & - & - & -  \\
    \hline
    $RNN_{1}:$ attn-pool\cite{jurczyk2016selqa} & - & - & - & - & - & - & .8643 & .8759 & - \\
    \bottomrule
  \end{tabu}
\end{table}

\begin{table}[H]
\caption{Experimental Results for the WikiQA dataset.}
  \label{wikiqa-results}
  \centering
  \begin{tabu} to \textwidth {lXXXXXXXXX}
    \toprule
    
    \multicolumn{1}{c}{Model} & \multicolumn{3}{c}{10 labeled instances} & \multicolumn{3}{c}{50 labeled instances} & \multicolumn{3}{c}{Full dataset}\\
    \cmidrule{2-10}
                        & MAP & MRR & NDCG & MAP & MRR & NDCG & MAP & MRR & NDCG \\
    \midrule
    CNN$_{hinge\_loss}$  & 0.5447 & 0.5577 & 0.6575 & 0.5919 & 0.6071 & 0.6942 & 0.6511 & 0.6669 & 0.7402\\
    CNN-DAN             & 0.5437 & 0.5582 & 0.6566 & 0.6047 & 0.6201 & 0.7042 & \textbf{0.6663} & \textbf{0.6822} & \textbf{0.7516} \\
    CNN-DAN$_{unlab.}$  & \textbf{0.5927} & \textbf{0.6068} & \textbf{0.6945} & \textbf{0.6127} & \textbf{0.6274} & \textbf{0.7104} & - & - & -  \\
    \hline 
    \citet{yang2015wikiqa} & - & - & - & - & - & - & 0.6520 & 0.6652 & - \\
    \citet{yin2015abcnn} & - & - & - & - & - & - & 0.6600 & 0.6770 & - \\
    \citet{dos2016attentive} & - & - & - & - & - & - & 0.6886 & 0.6957 & - \\
    \citet{wang2017bilateral} & - & - & - & - & - & - & \textbf{0.7180} & \textbf{0.7310} & - \\
    \bottomrule
  \end{tabu}
\end{table}

\begin{figure}[H]
\includegraphics[width=0.5\linewidth]{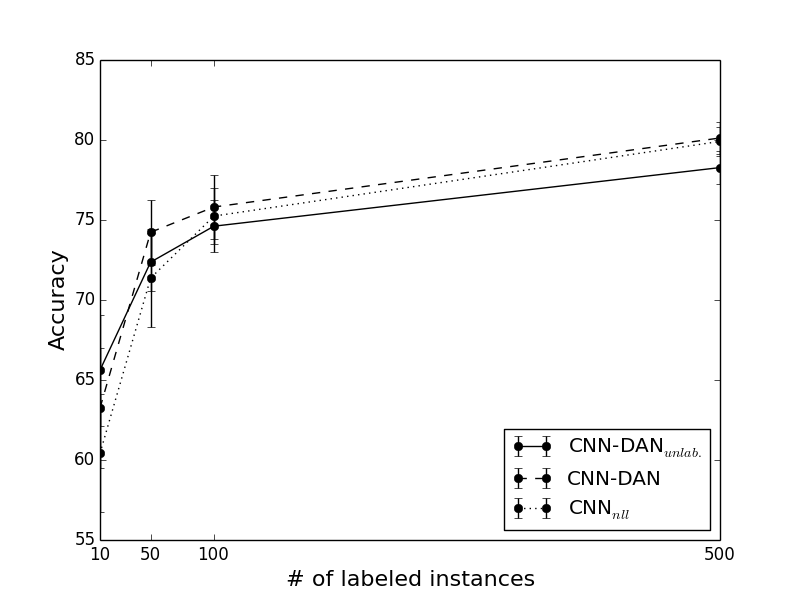}
   \centering
\caption{SSTB2}
\label{fig:sstb2}
\end{figure}
\vspace{-6ex}

\subsection{Text classification Results}
In Table \ref{sstb-results} and Fig. \ref{fig:sstb2} we present the experimental results for SSTB2 dataset.
We use accuracy to assess the output of the models.
We present results for CNN-DAN, CNN-DAN$_{unlab.}$ (that uses unlabeled data in $P$) and CNN$_{nll}$, 
which is the same CNN-based architecture of the predictor $P$ in our DAN for text classification (Fig. \ref{fig:pj-text}), 
but that is trained using the negative log likelihood loss function instead of the DAN framework.
Similar to the answer selection task, 
we present detailed results for datasets containing a different number of labeled instances: 10, 50 and full dataset.
Fig. \ref{fig:sstb2} also presents the accuracy for datasets with 100 and 500 labeled instances.

We can see in Fig. \ref{fig:sstb2} that, like in the answer selection task, the semi-supervised DAN (CNN-DAN$_{unlab.}$) gives a significant boost in performance when using only 10 labeled instances.
However, differently from the answer selection task, CNN-DAN$_{unlab.}$ is not able to improve upon CNN$_{nll}$ when we have more than 50 labeled instances. We believe this is mainly due to the difficulties of training the DAN with an unbalanced number of labeled and unlabeled instances. DANs for answer selection have proven to be more stable/easier to train probably because each instance in that task includes much more information (a question and a list of candidate answers) than in the sentiment classification case (a single sentence). We believe that additional hyperparameter tuning and perhaps some tricks to stabilize the min-max game would help to improve the performance of CNN-DAN$_{unlab.}$.

On the other hand, CNN-DAN, that does not used unlabeled data, was more stable for this task. 
CNN-DAN produced better results than CNN$_{nll}$ for all labeled set sizes, including the full dataset setup. 
Which means that, specially for small labeled set sizes, DAN was able to learning a loss function that is more effective than the negative log likelihood, which is probably the most widely used loss function for classification tasks. 

When compared to state-of-the-art results, we have better results than \citet{santos2014:coling}, 
who use an architecture that is very similar to our predictor.  Our results are comparable to the results of \citet{kim2014convolutional}, that uses a multi-channel CNN. However, our results are behind the results reported by \citet{hu2016deep}, that uses a more complex predictor architecture that leverages external knowledge.

\begin{table}[H]
  \caption{Experimental Results for the SSTB2 dataset.}
  \label{sstb-results}
  \centering
  \begin{tabular}{lccc}
    \toprule
    
    \multicolumn{1}{c}{Model} & \multicolumn{1}{c}{10 instances} & \multicolumn{1}{c}{50 instances} & \multicolumn{1}{c}{Full dataset}\\
    \cmidrule{2-4}
                        & Average Accuracy & Average Accuracy & Average Accuracy \\
    \midrule
    CNN$_{nll}$ & 60.42 $\pm$ (3.68) & 71.36 $\pm$ (3.04) & 84.38 $\pm$ (0.50) \\
    CNN-DAN     & 63.25 $\pm$ (3.78) & \textbf{74.23 $\pm$ (2.02)} & \textbf{84.70 $\pm$ (0.41)}  \\
    CNN-DAN$_{unlab.}$  & \textbf{65.62 $\pm$ (3.46)} & 72.37 $\pm$ (1.79)  & -  \\
    \hline
    \citet{santos2014:coling} & - & - & 82.0 \\
    \citet{kim2014convolutional}  & - & - & 86.6 \\
    \citet{hu2016deep}  & - & - & \textbf{89.4} \\
    \bottomrule
  \end{tabular}
\end{table}

\section{Conclusion}
\label{sec:conclusion}
Our experimental results evidence that DAN is a promising framework for both semi-supervised learning and learning loss functions for predictors. Going forward, we believe that improvements in training stability will bring additional gains for DAN. We tried some tricks used for training GANs such as minibatch discrimination and feature matching \cite{salimans2016improved}, but they did not help much for the two architecture presented in this paper.
Another research direction is on developing theoretical grounding for DANs with the focus on both semi-supervised and loss function learning. 



\bibliographystyle{abbrvnat} 
\bibliography{nips_2017} 

\end{document}